\documentclass[11pt]{article}
\usepackage{coling2020}
\usepackage{times}
\usepackage{url}
\usepackage{latexsym}
\usepackage{times}
\usepackage{soul}
\usepackage{amsfonts}
\usepackage{amsmath}
\usepackage{graphicx}
\usepackage{latexsym}

\usepackage{tikz}
\usepackage{pifont}
\newcommand{\cmark}{\ding{51}}%
\newcommand{\xmark}{\ding{55}}%
\usepackage[disable,textsize=tiny,linecolor=orange,backgroundcolor=yellow,textwidth=4em]{todonotes}
\usepackage{mnotes} 
\usepackage{soul}

\colingfinalcopy % Uncomment this line for the final submission

\title{AttViz: Online exploration of self-attention \\ for transparent neural language modeling}
\author{Bla\v{z} \v{S}krlj \\
  Jo\v{z}ef Stefan Institute \\
  Jo\v{z}ef Stefan International \\ Postgraduate School \\ \\
  \textbf{Shane Sheehan} \\
  University of \\
  Edinburgh  \And
  \textbf{Nika Er\v{z}en} \\
Jo\v{z}ef Stefan Institute\\\\\\\\
  \textbf{Marko Robnik-\v{S}ikonja} \\
  University of Ljubljana \And
  \textbf{Saturnino Luz} \\
  University of\\ Edinburgh \\\\\\
  \textbf{Senja Pollak} \\
  Jo\v{z}ef Stefan Institute
}

\date{}

\begin{document}
\maketitle
\begin{abstract}
Neural language models are becoming the prevailing methodology for the tasks of query answering, text classification, disambiguation, completion and translation. Commonly comprised of hundreds of millions of parameters, these neural network models offer state-of-the-art performance at the cost of interpretability; humans are no longer capable of tracing and understanding how decisions are being made. The attention mechanism, introduced initially for the task of translation, has been successfully adopted for other language-related tasks. We propose AttViz, an online toolkit for exploration of self-attention---real values associated with individual text tokens. We show how existing deep learning pipelines can produce outputs suitable for AttViz, offering novel visualizations of the attention heads and their aggregations with minimal effort, online. We show on examples of news segments how the proposed system can be used to inspect and potentially better understand what a model has learned (or emphasized).
\end{abstract}

\section{Introduction}
\label{sec:intro}
%\hl{MRS: Is AttViz a suitable name. The paper does not really visualize embeddings, but the self-attention mechanism. BS: The name stays as we already have a subdomain registered to AttViz. SP: If you think the name is not appropriate, it is possible to have really the AttViz as a domain for different visualization of EMBEDDIA project and AttViz/selfat or similar as a subdomain here? BS: If we update the tool with the actual embedding visualization, we would need to re-name again}
Contemporary machine learning that addresses text-related tasks adheres to the use of large \emph{language models}---deep neural network architectures that have gone through extensive unsupervised pre-training in order to capture context-dependent meaning of individual tokens \cite{devlin2019bert,liu2019roberta,yang2019xlnet}. Even though pre-training of such multi-million parameter neural networks can be expensive \cite{radford2019language}, many pre-trained models have been made freely available to the wider research community, unveiling the opportunity for the exploration of how, and why such large models perform well.
One of the main problems with neural language models is their \emph{interpretability}. Even though the models learn the task well (even at super-human levels), understanding the reasons for predictions and inspection of whether the models picked up irrelevant biases or spurious correlations can be a non-trivial task.

Approaches to understanding black-box (non-interpretable) neural network models often resort to \emph{post-hoc} approximations, e.g., SHAP \cite{lundberg2017unified}, and similar are not necessary internal to the model itself. 
%\hl{MRS: In the previous sentence> There are many approaches where explanation is learned together with the main task. The statement shall be modified. BS: added neural network models and necessary + the word often, which implies this is not always the case }
A potential way of extracting the token relevance is the attention mechanism \cite{bahdanau;2014,luong2015effective}. The attention mechanism learns token pair-value mappings, potentially encoding \emph{relations} between token pairs.
%\hl{MRS: The attention mechanism shall be explained here in layman terms.} 
When inspected as self-relations, the attention of a token w.r.t. itself (the diagonal element of the token attention matrix) potentially offers some insight into the importance of that token. Similar findings were also recently discussed when considering tabular data \cite{arik2019tabnet}.
%\hl{if possible, elaborate a bit more the previous sentence BS: Added some clarification on what this is} 
However, analytically, as well as numerically, exploration of attention can be a cumbersome task, resulting in the rise of approaches aimed at \emph{attention visualization}.
Visualization of (latent) embedding spaces is becoming ubiquitous in contemporary machine learning. For example, the Google's Embedding Projector\footnote{\url{https://projector.tensorflow.org/}} has offered numerous  visualizations for non-savvy users, making embedding projections to low dimensional (human-understandable) vector spaces simple and available \emph{online}. Even though visualization of simple embedding spaces is already accessible, visualization of complex neural network models' interior representations distributed across multiple embeddings (e.g., attention vectors), however, can be a challenging task.
%\hl{MRS: Previous sentence: what can be jointly coupled? Don't understand. BS: Changed it to mention attention vectors}
The works of \cite{liu2018visual} and \cite{8614007} are examples of attempts at unveiling the workings of black-box  attention layers and offering an interface for human researches to learn and inspect their models.%\hl{check BS: okay, yes}
%SP:I reformulated the sentence, as in next section you mention older works. OLD We acknowledge the works of \cite{liu2018visual,8614007} as the initial attempts at unveiling the workings of black-box  attention layers and offering an interface for human researches to learn and inspect their models. For example,
\cite{liu2018visual} visualize %\hl{token pair-value mappings}%SP: to se cudno slisi: token pair-value pairs - potem morda rajsi token pair-value mappings, kot imas zgoraj uvedeno, pa se ni cisto jasno, kaj pa kar e attention values of the token pairs BS: Okej, mappings je tudi smiselno v tem kontekstu}
, as well as offer possible coloring of the attention space. 
%\hl{MRS: A reader does not know of key-value pairs at this point.BS: This is now introduced at the beginning of the section as token pair-value mappings}
Further, \cite{8614007} visualized self-attention with examples in sentiment analysis. The main contributions of AttViz are multi-fold, and can be stated as follows.
AttViz focuses exclusively on self-attention and introduces two novel ways of visualizing this property while being available online and accessible to a wider audience. AttViz can interactively aggregate the attention vectors and offers simultaneous exploration of the output probability space, as well as the attention space.

%\hl{MRS: Previous sentence: this is not well stated - it implies that the main novelty of the AttViz is the online character of the tool and potentially its speed. The contributions shall be clearly stated here. We shall introduce the name AttViz, describe what it is, its main properties and the scope where it works. BS: Re-stated the claims}

The remainder of this work is structured as follows. In Section~\ref{sec:attentionviz}, we discuss the works, related to the proposed AttViz approach. In Section~\ref{sec:AttViz}, we present the key ideas and technical implementation of AttViz, followed by our use case -- a study of news segments in Section~\ref{sec:usage}. Finally, we discuss (in Section~\ref{sec:discussion}) the overall capabilities of AttViz.

\section{Attention visualization}
\label{sec:attentionviz}
Visualization of the attention mechanism for text has recently emerged as an active research area due to an increased popularity of attention based methods in natural language processing.  Recent deep neural network language models such as BERT \cite{devlin2019bert}, XLNet \cite{yang2019xlnet}, and RoBERTa \cite{liu2019roberta} are comprised of multiple \emph{attention heads}---separate weight spaces each associated with the input sequence in \emph{a unique way}. Language models consist of multiple attention matrices, all contributing to the final prediction. Visualising the attention weights from each of attention matrix is thus an important component in understanding and interpreting these models. 

%\hl{MRS: This description of attention heads is too abstract. I suggest that you illustrate an example of attention heads for a simple input in the introduction. BS: Changed to matrices directly, no need for heads at this point indeed}

The attention mechanism which originated in the work on neural machine translation lends
itself naturally to visualisation. \cite{bahdanau;2014}   used \emph{heat maps} to display the attention weights between input and output text. This visualisation technique was first applied to the task of translation but found use in many other tasks such as visualising an input sentence and output summarization \cite{rush:2015} and visualizing an input document and textual entailment hypothesis \cite{Rocktschel:2015}. In these heat map visualisations, a matrix or a vector is used to represent the learned alignments and color intensity illustrates attention weights. This provides a summary of the attention patterns describing how they map the input to the output. For classification tasks, a similar visualisation approach can be used to display the attention weights between the classified document and the predicted label \cite{yang:2016,Tsaptsinos:2017}. Here, the visualisation  of attention  often displays the input document  with the attention weights \emph{superimposed} onto individual words. The superimposed attention weights are represented similarly to heat map visualisations using a color saturation to encode attention value.

 An alternative visual encoding of attention weights is a bipartite graph visualisation. Here attention weights are represented by edge weights or thickness between two lists of words. This technique has been used to help interpret model output in neural machine translation \cite{lee:2017}, in natural language inference \cite{liu:2018}, and for model debugging \cite{Strobelt:2018}. A version of this visualisation which was designed specifically for multi-head self-attention \cite{Vaswani:2017} uses color hue to encode the attention head associated with each weight. The color is applied to the edges and is superimposed as a color strip over the node words. The intensity of the colors in the strips at each word position summarises the distribution of attention weight for that word across the heads. This multi-head self visualisation technique was recently extended \cite{vig:2019} with two visualisations. The first, called ``Model View'', is a 
% \hl{MRS: What multiples means here? BS: Removed this explanation alltogether} 
 visualisation of the bipartite graphs for each layer and head in the system. The second visualization,  ``Neuron View'', drills down to the computation of the attention score associated with each weighed edge in the bipartite graph. The element wise product, dot product, and softmax values are all visualised using coloured elements with saturation representing the magnitude of the value. 
 This visualisation provides some insight into how each attention weight was computed while still providing the overview of attention weights.

 The purpose of the proposed AttViz is to unveil the attention layer space to human explorers in an intuitive manner. The tool emphasizes \emph{self-attention}, that is, the diagonal of the token-token attention matrix which possibly corresponds to \emph{relevance} of individual tokens. By making use of alternative encoding techniques, the attention weights across the layers and heads can be explored dynamically to investigate the interactions between the model and the input data.
The proposed AttViz differentiates from existing visualization tools as follows. The focus of the tool is, as stated, self-attention, implying visualization of (attention-annotated) input token sequences can be carried out directly. We developed a novel visualization technique, where self-attention values are on per-token basis visualized across the input sequence for each self-attention vector. Further, AttViz offers visualization of the distribution of the attention values across the token sequence along with relevant aggregations, such as the min/max and similar.  Finally, the tool simultaneously shows both the prediction probabilities, making interpretation of the self-attention space even more transparent, and with it the information on potential alternative classifications.
%\hl{MRS: We shall contrast AttViz with the describes visualizations. BS: added a sentence with explanation}

\section{AttViz: An online toolkit for visualization of self-attention}
\label{sec:AttViz}
We built AttViz, an online solution that can be coupled with existing language models from the PyTorch-transformers library\footnote{\url{https://github.com/huggingface/transformers}}---one of the most widely used resources for language modeling. The idea behind AttViz is that it is  \emph{lightweight}, as it does not offer (online) neural model training, but facilitates the exploration of \emph{trained} models. Along with AttViz, we provide a set of Python scripts that take as an input a trained neural language model and output a JSON file to be used by AttViz visualisation tool. %comprised of the information the AttViz can visualize is offered. 
A common pipeline for using AttViz is as follows. First, a transformer-based trained neural network model is used to obtain predictions on a desired set of instances (texts). The predictions are converted into the JSON format, suitable for AttViz, along with the attention space of the language model. The JSON file is loaded into AttViz (on the user's machine client side), where its visualization and exploration is possible. We next discuss the proposed visualization of the self-attention.

\subsection{Visualization of self-attention}
In this section, we discuss the proposed visualization schemes that emphasize different aspects of self-attention. 
%\hl{MRS: I suggest that we couple this description with references to one of the figures. BS: added descriptions where needed} 
The initial AttViz view offers sequence-level visualization, where each (byte-pair encoded) token is equipped with a self-attention value based on a given attention head (see Figure~\ref{fig:pol11}; central text space). Following the first row that represents the input text, consequent rows correspond to attention values that  represent the importance of a given token with respect to a given attention head. As discussed in the empirical part of this paper (Section~\ref{sec:usage}), the rationale for this display is that commonly, only a certain number of attention heads are activated (colored fields), thus visualization must entail both the whole attention space, as well as emphasize individual heads (and tokens).

The same document can also be viewed in the ``aggregation'' mode (Figure~\ref{fig:pol12}), where the attention sequence is shown across the token space. The user can interactively explore how the self-attention varies for individual input tokens, by changing both the scale, as well as the type of the aggregation used, the visualization can be used to emphasize various aspects of the self-attention space.

%hl{MRS: Refer to elements shown in the figure. BS: added reference + additional description}
The second developed visualization (Figure~\ref{fig:pol12}) is the overall distribution of attention values across the whole token space. Resembling a time series, for each consequent token, the attention values are plotted separately. This visualization offers an insight into \emph{self-attention peaks}, i.e. parts of the attention space focused around certain tokens that potentially impact the performance and the decision making process of a given neural network. This view also emphasizes different aggregations of the attention vector space for a single token (e.g., mean, entropy, and maximum). The visualization, apart from the mean self-attention (per token), also offers the information on maximum and minimum attention values (red dots), as well as the remainder of the self-attention values (gray dots). The user  can this way explore both the self-attention peaks, as well as the overall spread.

%\hl{MRS: Above three visualizations are mentioned, here we refer to two? } The two proposed visualization views can be used interchangeably for better usability of the proposed tool. Even though the visualization of \emph{raw} attention values  is useful, we next discuss the \emph{attention aggregates} implemented as part of AttViz.

\subsection{Aggregation of self-attention}
\label{sec:agg}
\begin{figure*}[h!]
    \centering
    \includegraphics[width = \linewidth]{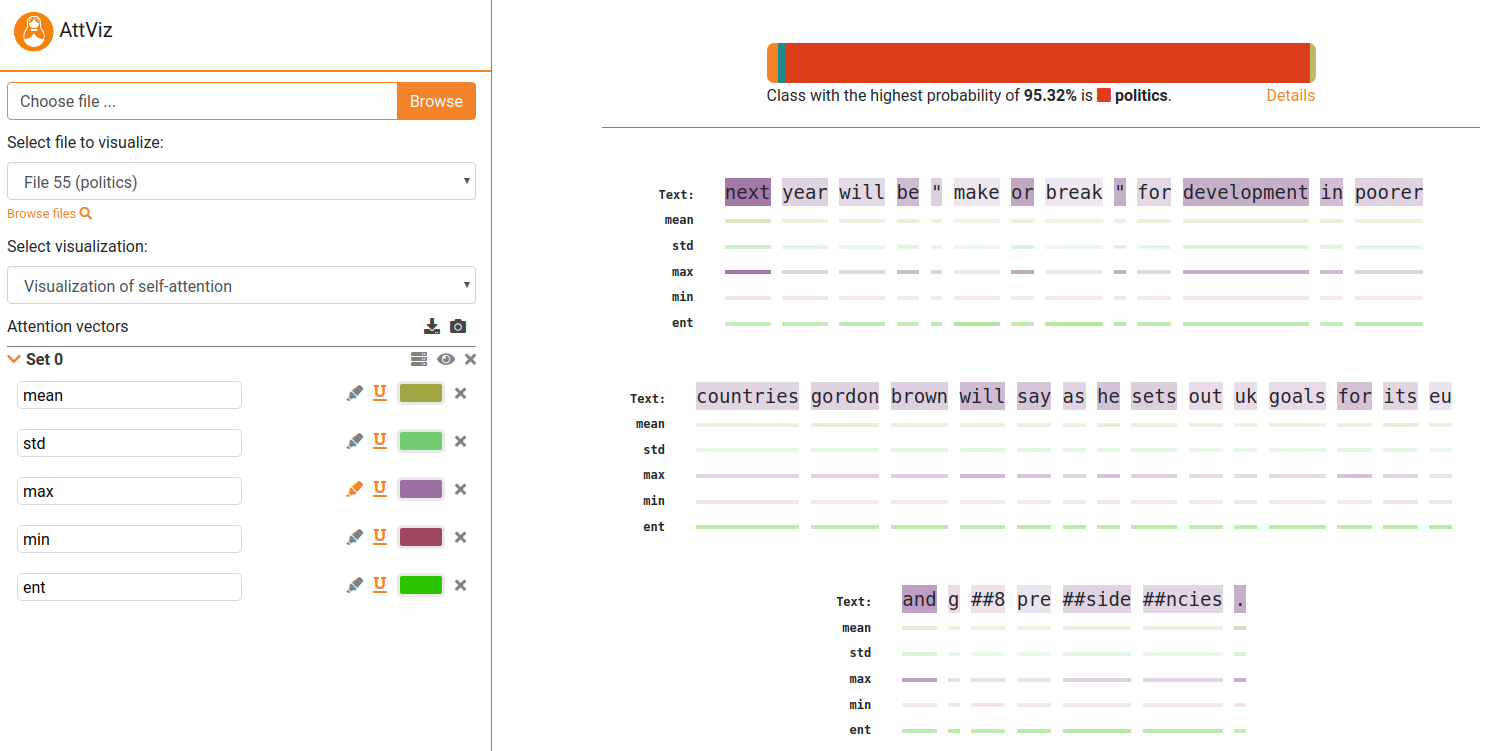}
    \caption{Visualization of aggregations. The document was classified as a politics-related topic, it can be observed that aggregations emphasize tokens such as ``development'',``uk'' and ``poorer''. The user can also highlight desired head information -- in this example the maximum attention (purple) is highlighted.}
     %   \caption{Visualization of the attention vectors for input byte pair encoded space. The last six rows correspond to various aggregation schemes implemented as part of AttViz. The maximum aggregation, for example, emphasizes tokens such as ``civil'' and  ``servants'' and similar. The figure demonstrates the situation observed throughout the space of all considered documents---only a few heads are ``active'', whilst the remainder does not appear to impact a given prediction.}
      %  \hl{SS: I don't think its clear which row you are referring to as "maximum aggregation". Civil appears to be most strongly emphasised in rows 2 and 3. not in the max agg row? The aggregations seem to be out biggest novelty(from the previous section) and should be emphasised more in the examples}
    %\hl{MRS: Similarly as above. The bottom tokens are strange. Can you put them on the top line, together with other tokens?}}
    \label{fig:pol11}
    \includegraphics[width = \linewidth]{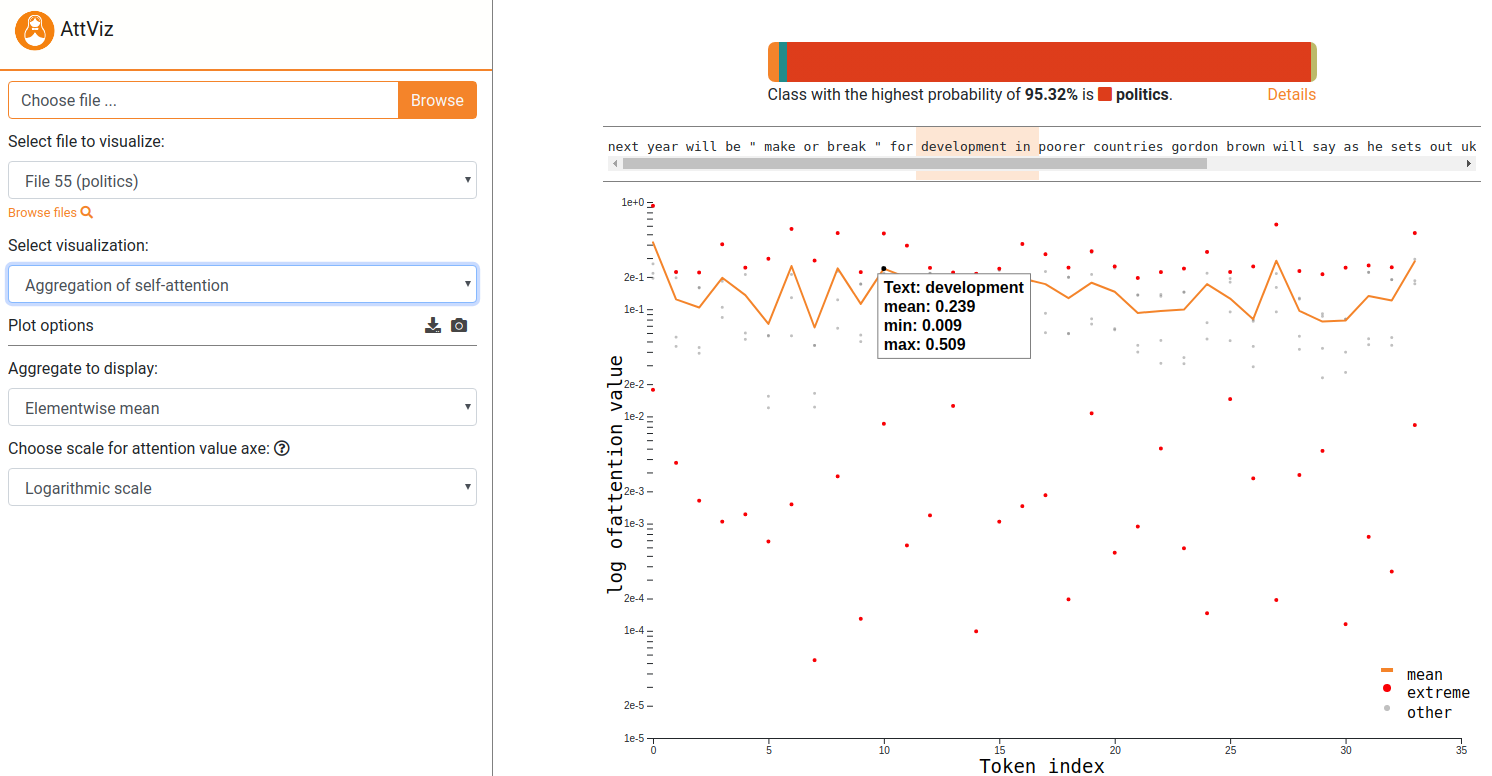}
    \caption{The interactive series view. The user can, by hoovering over the desired part of the sequence, inspect the attention values and their aggregations. The text above the visualization is also highlighted automatically.}
   % \caption{Visualization of the attention vector with dynamic highlight option (orange)\hl{SS: The yellow highlighted section is missing?}. The user can interactively traverse the attention sequence by hovering over with mouse pointer and monitor the corresponding token space. The highlighted region (orange) represents the current part of the input sequence that is being inspected---for a given token, the attention distribution across all heads is shown.
  %  The shown view displays attention values as gray dots, where mean (orange), maximum and minimum (red) values are colored for easier inspection.
    %The uppermost bar visualization shows probabilities for assigning the observed document into one of the possible classes---the most probable one is politics (red).}
    %\hl{Improve the explanation, it is not sufficiently clear; Also, in the figure there is a space missing in y label- log of SPACE attention value.}
    %\hl{MRS: The figure need better explanation here and in the text. What are the dots? What different attention heads show? What was classsified? What the line shows? How this can be interpreted? }}
     \label{fig:pol12}
\end{figure*}

We apply several aggregation schemes  across the space of individual tokens.
Consider a matrix $A \in \mathbb{R}^{h \times t}$, where $h$ is the number of attention vectors and $t$ the number of tokens. We consider various aggregations across the second dimension of the attention matrix $A$ (index $j$).
In entropy based calculation, we denote with $P_{ij}$ the probability of observing $A_{ij}$ in the $j$-th column. The $m_j$ corresponds to the number of unique values in that column.
The proposed schemes are summarized in Table~\ref{tab:agg}.
\begin{table}[h]
    \centering
        \caption{Aggregation schemes used in AttViz.}
    \begin{tabular}{c|c}
\hline
       Aggregate name  & Definition \\ \hline
        Mean(j) (mean)& $\frac{1}{h}\sum_i A_{ij}$ \\ 
        Entropy(j) (ent)& $-\frac{1}{m_j}\sum_{i=0}^{h} P_{ij}\log P_{ij}$ \\
        Standard deviation(j) (std)& $\sqrt{\frac{1}{h - 1}\sum_i (A_{ij} - \overline{A_{ij}})^2}$ \\
        Elementwise Max(j) (max)& $\max\limits_{i}(A_{ij})$\\
        Elementwise Min(j) (min)& $\min\limits_{i}(A_{ij})$ \\ \hline
%        Elementwise Max - Min (mmin)& $\max\limits_{i}(A_{ij}) - \min\limits_{i}(A_{ij})$ \\ \hline
    \end{tabular}
    \label{tab:agg}
\end{table}
The attention aggregates can also be visualized as part of the aggregate view (Figure~\ref{fig:pol12}), where, for example, the mean attention is plotted as a line along with the attention space for each token, depicting the \emph{dispersion} around certain parts of the input text. %\hl{MRS: Missing references to figures and its parts.}

\section{Comparison with state-of-the-art}
\label{sec:sota-comparison}
In the following section we discuss in more detail the similarities and differences of AttViz with other state-of-the-art visualization approaches. Comparisons are summarized in Table~\ref{tab:comparison}. The neat-vison package is available at\footnote{\url{https://github.com/cbaziotis/neat-vision}}.
\begin{table}[h]
    \centering
    \resizebox{\textwidth}{!}{
    \begin{tabular}{c|cccc}
    Approach & AttViz (this work) & BertViz \cite{vig:2019} & neat-vision &  NCRF++ \cite{yang2018ncrf} \\ \hline
    Visualization types & sequence, aggregates & head, model, neuron  & sequence & sequence \\
    Open source & \cmark & \cmark & \cmark & \cmark \\
    Language & Python + Node.js & Python & Python + Node.js & Python \\
    Accessibility & Online & Jupyter notebooks & Online & script-based \\
    Sequence view & \cmark & \cmark & \cmark &\cmark \\
    Interactive & \cmark &\cmark &\cmark &\xmark \\
    Aggregated view & \cmark & \xmark & \xmark & \xmark \\
    Target probabilities & \cmark & \xmark & \cmark & \xmark \\
    Compatible with PyTorch Transformers? \cite{Wolf2019HuggingFacesTS} & \cmark & \cmark &\xmark & \xmark \\
     token-to-token attention & \xmark & \cmark &\xmark & \cmark
    \end{tabular}}
        \caption{Comparison of different aspects of the attention visualization approaches.}
%        \hl{SS, SP: Footnote 4 from the table caption does not appear in the output pdf}
        %\hl{SS: I increased table size as it was difficult to read}
        %\hl{SS:I like the idea of comparing the tools capabilities with those of other state of the art. DO we need to also mention things the others can do the attViz can't? As far as I understand, as mentioned in the example (please correct me if I'm wrong), AttViz does not visualise token to token attention patterns/matrices, While Vig or an attention heatmap  do(eg https://github.com/zhaocq-nlp/Attention-Visualization). It may be good to have clarified why we dont focus on or include these BS: Indeed, I've added this line + some clarification}
    \label{tab:comparison}
\end{table}

The main novelties introduced as part of AttViz are the capability to aggregate the attention vectors with four different aggregation schemes, offering insights both into the average attention but also its dispersity around a given token. The neat-vision project is the closest to AttViz's functionality, with the following differences. It is not directly bound to PyTorch transformers library, requiring additional pre-processing on the user-side. Similarly, the fast switching between the sequence and aggregate view are more emphasized in AttViz, as they offer more general overview of the attention space. The class probabilities are to our knowledge available in both tools, offering simultaneous exploration of both input and output space at the same time.
\section{Example usage: News visualization}
\label{sec:usage}
In this section, we present a step-by-step use of the server along with potential insights the user can obtain.
\begin{figure*}[!t]
    \centering
    \includegraphics[width = \linewidth]{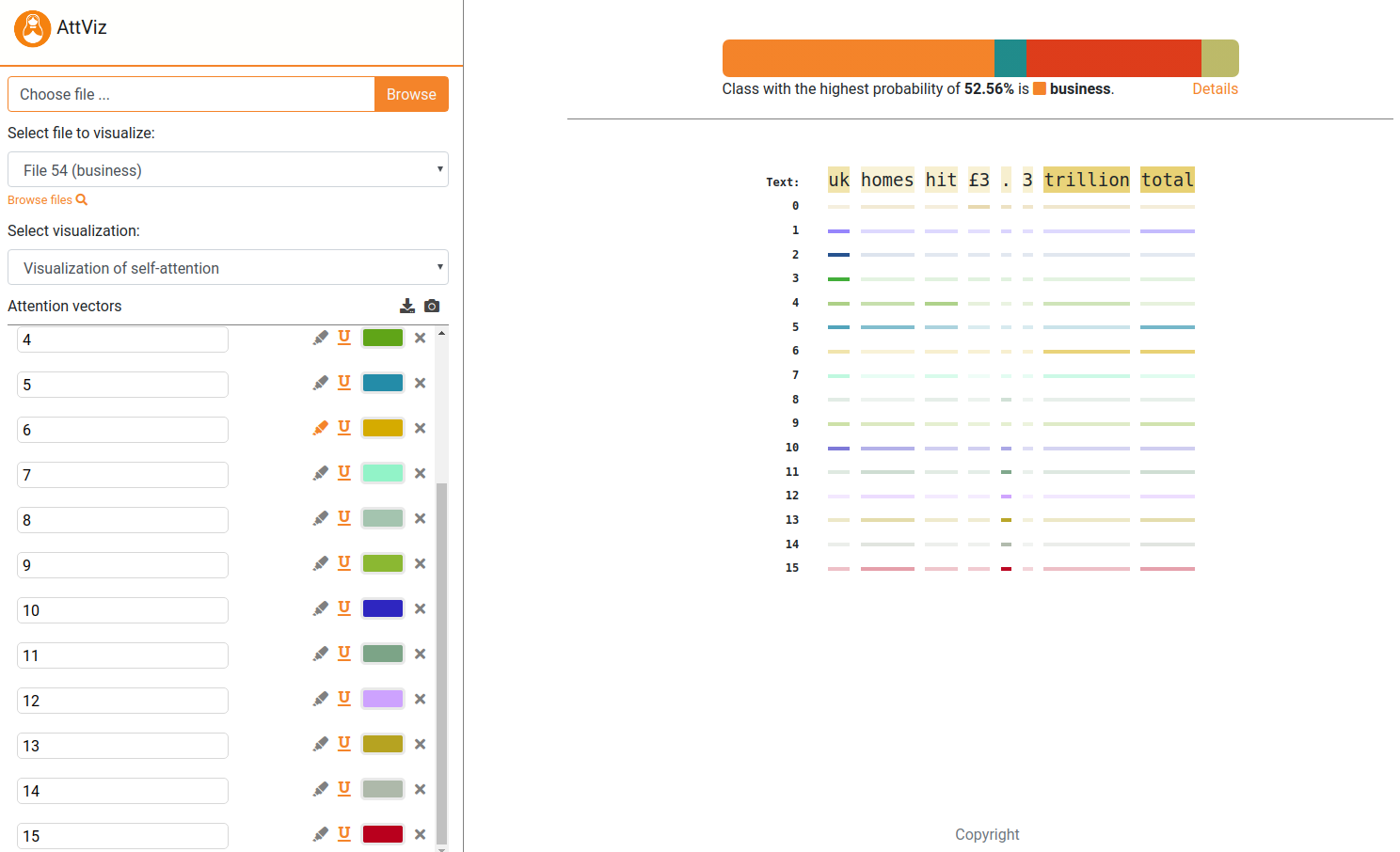}
    \caption{Visualization of all attention heads. The sixth heads's self attention is also used to highlight the text. The document was classified as a business-related, which can be linked to high self attention at the ``trillion'' and ``uk'' tokens. Note also that, compared to the first two examples (Figures 1 and 2), the network is less \emph{certain} -- the business and politics classes were predicted with similar probabilities (orange and red parts of the bar above visualized text).}
  %  \caption{A document (correctly) classified as a business topic. Note that heads 5-7 and 10 identify strong signal at the token ``economy'', which arguably corresponds to the correctly predicted topic (business). Several attention heads highlight also the first token ``giant''.  }
    %Note that the maximum aggregation (red vector) also emphasizes the same tokens.\hl{Meni pri agregirani verziji economy ne izgleda nic bolj poudarjen od skoraj vseh drugih? Morda das zadnji stavek ven?}}
   % \hl{SS: Do 4,9,11 and 12 not also identify strong signal at the token ``economy''? }
     
         %\hl{This may not be a useful comment for the paper, but for quantitative visual comparisons color saturation is a very low ranking variable, meaning it is difficult to say by how much one value is greater than another if the data is mapped to color saturation. Also, using different colors makes the comparison across rows even more difficult as some are naturally perceived as more saturated. Maybe for version 2.0 we can look at some different options for the renderings BS: Indeed, you are right. This is just the first step.}
    \label{fig:bus}
\end{figure*}
The examples are based on the BBC news data set\footnote{\url{https://github.com/suraj-deshmukh/BBC-Dataset-News-Classification/blob/master/dataset/dataset.csv}}  \cite{greene06icml}  that contains 2,225 news articles on five different topics (business, entertainment, politics, sport, tech). The documents from the dataset were split into short \emph{segments}. %by using the double new line character.
The splits allow easier training (manageable sequence lengths), as well as easier inspection of the models. We split the dataset into 60\% of the documents that were used to train a BERT-base \cite{devlin2019bert} model, 20\% for validation and 20\%  for testing\footnote{The obtained model classified the \emph{whole} documents into five categories with 96\% Accuracy, which is comparable with the state-of-the-art performance \cite{Trieu:2017:NCS:3155133.3155206}; however, note that the train, validation, and test splits were randomly created. For prediction and visualisation, only short segments are used}.
%\hl{is the State of the art also done on short segments or is it document level?BS: the model was trained on the whole documents, segments were used for prediction here.}

%\hl{specify if this corresponds to the original train test split BS: added the claim} 
%In the remainder of this work, we consider only correctly classified examples from the testing set.
%\hl{Morda ta stavek ni potreben in rajsi pri prvem primeru, ko ga navedes, reces da je example of a correctly classified instance from the test  oz. mislim, da to itak poves?BS: to je res ja, ampak a skodi?}
%\hl{Should you mention the performance of the model? BS: added as a footnote}
The fine-tuning of the BERT model was conducted as discussed in the examples of the PyTorch-Transformers library \cite{Wolf2019HuggingFacesTS}. The best-performing hyper parameter combination was using 3 epochs with the sequence length of 512 (other hyper parameters were left at their default values). We used  Nvidia Tesla V100 GPU processor for fine-tuning. While more recent larger language models such as e.g., XLNet  \cite{yang2019xlnet} could produce better accuracy, the idea and the use of AttViz visualizations is the same; hence, we selected the most commonly used model (BERT-base).

The main user interface of AttViz is displayed in Figures~\ref{fig:pol11} and ~\ref{fig:pol12} and ~\ref{fig:bus}. In the first example (Figure~\ref{fig:pol11}), the user can observe the main view that consists of two parts. The leftmost part shows (by id) individual self-attention vectors, along with visualization, aggregation and file selection options. The file selection indexes all examples contained in the input (JSON) file. Attention vectors can be colored with custom colors, as shown in the central (token-value view). The user can observe that, for example, the violet attention head (no. 5) is active, and emphasizes tokens such as ``servants'' (from civil servants), which indicates %are indeed 
a politics-related topic (as correctly classified).
%\hl{servants is clearly most marked, official documents are less marked than e.g. the previous words? maybe keep only tokens, such as ``servants'' (from civil servants) BS: That is a good suggestion, kept only servants.}
%\hl{MRS: I see no colors in the rightmost part of Figure 1. Better explain individual elements of the image. BS: Changed the sentence}
%\hl{MRS: Better point out the attention vectors on Figure 1. Can you relate them to the chosen text and classification? At the moment, the reader cannot see why these visualizations are useful. BS: added the immediate observation, this will potentially be replaced by a new example.}
Here, the token (byte-pair encoded) space is shown along with self-attention values for each token. The attention vectors are shown below the token space and aligned for direct inspection (and correspondence). Further, the upmost visualization in the right part of the view shows probabilities (obtained via softmax normalization of the output layer weights) of the considered document belonging to a given class. This functionality was added to help human explorers investigate the correspondence between the actual classification and the classified text. Using ``Details'' section below the probability legend, the distributions across the class space can be further inspected. %\hl{MRS: Figure 2 is not explained and referenced in the text. BS: Added the reference and the explanation}

In Figure~\ref{fig:pol12}, the user can observe the same text segment as an attention series spanning the input token space.
%\hl{There is a mistake here, Figure 3 is giant waves damage example, however here you still relate to Figure 1 as before - please check the last 2-3 paragraphs, and merge it. Introduce also Figure 2. For the same example, Figure 2 shows....} 
Again, note that tokens, such as ``trillion'' and ``uk'' correspond to high values in a subset of the attention heads, indicating their potential importance for the obtained classification. However, we observed that only a few attention heads ``activate'' with respect to individual tokens, indicating that other attention heads are not focusing on the tokens themselves, but possibly on \emph{relations} between them. This is possible and the attention matrices contain such information, yet the study of token relations is not the focus of this work (see \cite{vig:2019} for such a visualization). In this work we focus on self-attention as such information can be directly mapped across token sequences, emphasizing tokens that are of relevance to the classification task at hand. Consequently, we see AttViz as being the most useful when exploring models used for classification of hatespeech or similar news texts, where individual tokens carry key information for classification.
 %\hl{SS: It may be good to have clarified why we don't focus on or include these other non diagonal sections of the attention matrix. Do we have a good motivation or do we even need one? Earlier we stated " The tool emphasizes self-attention, that is, the diagonal of the token-token attention
%matrix  which  possibly  corresponds  to relevance of  individual  tokens." Can we give some reason why token to token attention patterns are not relevant to the classification decision and can be ignored in our tool design? Perhaps it is obvious and would be clear to people with better understanding then I.}

In the example in Figure~\ref{fig:bus}, we visualize a short segment related to uk homes and spending. Note that the text is shown after the preprocessing consisting of byte-pair encoding and lower-casing.

The segment was correctly classified as business-related.
Tokens, such as ``trillion'', ``uk'' and ``total'' are all associated with high attention.
The example shows how different attention heads detect different aspects of the sentence, even at the single token (self-attention) level. The user can observe that the next most probable category for this topic was politics (red color), which is indeed a more sensible classification than e.g., sports. The example shows how interpretation of the attention can be coupled with the model's output for increased interpretability.

A careful inspection of the remainder of the documents revealed that in the majority of cases, the first token is also emphasized. We believe the following reasons can induce this observed bias.
First, as the BERT-base model was used for the classification task, the model was only fine-tuned on the news data set (for a few epochs), after being extensively pre-trained on vast amounts of text. The pre-training phase could introduce the bias, as the model is implicitly forced to learn to predict the next token, indicating that the first token in the classified segment will be of high ``relevance''.
%\hl{Meni ni cisto jasna povezava te argumentacije z dejstvom, da je ojacan first token - lahko morda se kaj dodatno pojasnis? Ce nic ne fine-tunam, potem je vse na zacetku pobarvano---zdi se, da pre-training gleda le kratkorocno} 
In the second interpretation, when the first token is a content work, it can already carry a lot of meaning for the whole sentence, thus it could be reasonably relevant for the task.
%\hl{Ampak zgoraj je giant v bistvu adjective. Morda When the first token is a content word (za razliko od function wordov, ki so v anglescini pogosto na prvem mestu?). A ni pa kake fore s prvim CLS toknom?  BS: Ja, to je se bolj smiselno}
%\hl{MRS: A more engaged use case which would better demonstrate the usefulness of the three visualizations would be welcome. I suggest that a really simple revealing example is added to Section 3 and illustrates descriptions of the method. Section 4 shall demonstrate practical usefulness. BS: Indeed, this will be added if time permits.}

\section{Critical overview of AttViz and Conclusions}
\label{sec:discussion}
As AttViz is an online toolkit for facilitated attention exploration, we discuss possible concerns regarding its usefulness. One of the main issues with online methods is privacy. Currently, AttViz does not employ any anonymization strategies, meaning  that private processing of the input data is not guaranteed. We believe that this issue can be addressed as a part of further work or with a private installation of the tool.
Further, AttViz leverages users' computing capabilities, meaning that too  large data sets can cause memory overheads (e.g., several millions of examples). We believe that such situations are difficult to address with AttViz, however, instances can be filtered prior to being used in AttViz. This would enable seamless exploration of a subset of the data (e.g., only (in)correctly predicted instances, or certain time slot of instances).
In terms of functionality, AttViz is focused on the exploration of \emph{self-attention}. We realize that the self-attention is not necessarily the only important aspect of a neural network that needs to be inspected, but it is possibly the one, where visualisation techniques have been the least  explored. Similarly to the work of \cite{liu2018visual}, we plan to further explore potentially interesting \emph{relations} emerging from the attention matrices.

Finally, we believe AttViz could be further extended with a larger database of popular models and a back-end functionality, enabling it to, e.g., fine-tune models.  Such endeavors are out of the scope of this paper---the current version of AttViz is lightweight, can be hosted by anyone (with minimal requirements overhead) and performs fast when considering exploration of self-attention.

%\hl{Manjka sekcija Conclusion, kjer vsaj na hitro povzames, kaj je narejeno.}

\section{Availability}
The tutorials and other input preparation scripts are available at: \url{https://github.com/SkBlaz/attviz}. The server is live at: http://attviz.ijs.si.

\section*{Acknowledgements}

Omitted for anonymity reasons.
%\hl{Blaz, in reference section: please substitute  ArXiv with non-preprint equivalents when available, at least: Yang et al.: https://papers.nips.cc/paper/8812-xlnet-generalized-autoregressive-pretraining-for-language-understanding.pdf.}

% include your own bib file like this:
\bibliographystyle{coling}
\bibliography{coling2020}

\end{document}